\documentclass[letterpaper, 10 pt, conference]{ieeeconf}  

\IEEEoverridecommandlockouts                              

\usepackage[utf8]{inputenc} 
\usepackage[T1]{fontenc}    
\usepackage{hyperref}       
\usepackage{url}            
\usepackage{booktabs}       
\usepackage{amsfonts}       
\usepackage{nicefrac}       
\usepackage{microtype}      
\usepackage{xcolor}         
\usepackage{wrapfig}        

\usepackage[ruled,vlined]{algorithm2e}
\usepackage[noend]{algpseudocode}
\SetKwInput{KwInput}{Input}                
\SetKwInput{KwOutput}{Output}              

\usepackage{times}
\usepackage{epsfig}
\usepackage{graphicx}
\usepackage{amsmath}
\usepackage{amssymb}
\usepackage{bbm}
\usepackage{microtype}
\usepackage{comment}

\makeatletter
\def\thickhline{%
  \noalign{\ifnum0=`}\fi\hrule \@height \thickarrayrulewidth \futurelet
   \reserved@a\@xthickhline}
\def\@xthickhline{\ifx\reserved@a\thickhline
               \vskip\doublerulesep
               \vskip-\thickarrayrulewidth
             \fi
      \ifnum0=`{\fi}}
\makeatother

\newlength{\thickarrayrulewidth}
\title{\LARGE \bf
Learning to Rearrange with Physics-Inspired Risk Awareness 
}

\author{ {Meng Song$^{1}$}\quad 
{Yuhan Liu$^{1}$} \quad 
{Zhengqin Li$^{2}$}\quad 
{Manmohan Chandraker$^{1}$}\quad \\[2mm]
{$^{1}$UC San Diego }\quad {$^{2}$Meta}
}

\begin{document}

\maketitle
\thispagestyle{empty}
\pagestyle{empty}

\begin{abstract}
\label{sec:abstract}
Real-world applications require a robot operating in the physical world with awareness of potential risks besides accomplishing the task. A large part of risky behaviors arises from interacting with objects in ignorance of affordance. To prevent the agent from making unsafe decisions, we propose to train a robotic agent by reinforcement learning to execute tasks with an awareness of physical properties such as mass and friction in an indoor environment. We achieve this through a novel physics-inspired reward function that encourages the agent to learn a policy discerning different masses and friction coefficients. We introduce two novel and challenging indoor rearrangement tasks -- the variable friction pushing task and the variable mass pushing task -- that allow evaluation of the learned policies in trading off performance and physics-inspired risk. Our results demonstrate that by equipping with the proposed reward, the agent is able to learn policies choosing the pushing targets or goal-reaching trajectories with minimum physical cost, which can be further utilized as a precaution to constrain the agent's behavior in a safety-critic environment.
\end{abstract}

\section{Introduction} \label{sec:intro}

There has been significant recent advances in applying reinforcement learning algorithms to a variety of domains ranging from playing games  \cite{alphago} \cite{dqn} to solving navigation and manipulation tasks in simulation environments \cite{igibson, ai2thor, habitat, sapien, robosuite, rearrangement}. However, in many real-world robotic problems, avoiding risky behaviors in deployment is as crucial as optimizing the task objective. There is a rich body of prior work on safe RL \cite{safe_rl} defining risk and enforcing safety from various perspectives, such as imposing constraints on state-action space \cite{safe_explore} and value functions \cite{CQL, learning_to_be_safe}, restricting the policies to be close to safe demonstrations \cite{apprentice_learn, safe_il}, etc. Motivated by the fact that many damage outcomes (e.g. falling down, dropping an object) are induced by the physical process of robot interacting with environments, we consider to model the risk in terms of physical cost over the trajectories. Besides collisions with the obstacles, physical properties such as the frictions of terrains and weights of the objects are also informative learning signals for the agent to assess the risk of its behaviors. However, these factors have not been sufficiently captured by the goal states defining the success of the classical navigation and manipulation tasks \cite{language_goal, habitat, sapien, metaworld, vision_goal_navigation}. 

\begin{figure}[!!t]
 \centering
 \includegraphics[width=1.0\columnwidth]{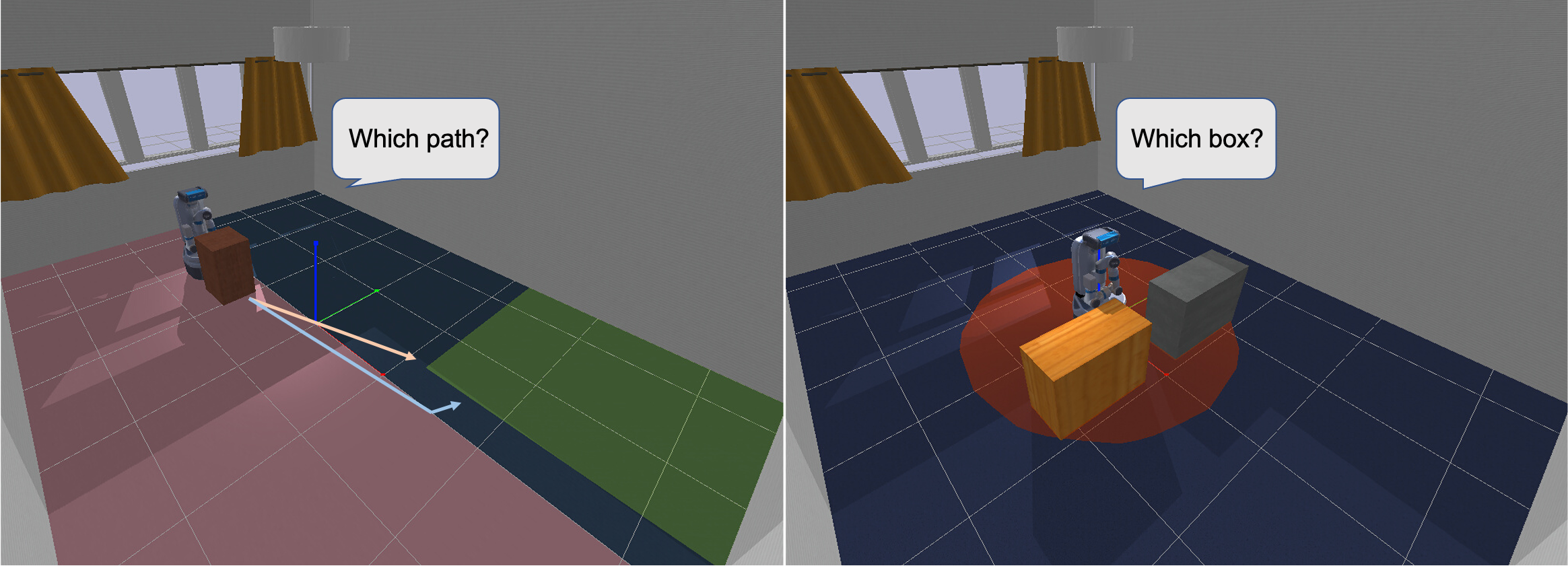}
 \caption{Left: The agent must learn to account for differing friction coefficients for the pink and blue sections of the floor, in order to efficiently push the box to the goal region (indicated in green). Right: The agent must learn to account for difference in masses for the orange and gray boxes, to expend the least physical cost in pushing  one of them outside the circular region.
 }
\label{fig:teaser}
\vspace{-0.5cm}
\end{figure}

To this end, we study how to incorporate physical cost into the reward function to guide the policy learning in indoor rearrangement tasks. We build two novel and challenging tasks featuring a wheeled robot pushing boxes to goal regions with physically variant properties (Fig. \ref{fig:teaser}). Accomplishing these tasks requires the agent to minimize its physical cost besides reaching the successful states. Our experiments have shown that the learned policies under this new reward function are able to stably distinguishing between different masses and frictions.

\section{Physical Cost} \label{sec:energy}
To estimate the physical cost in an arrangement task, we compute the amount of virtual physical work $E(t)$ the agent spends on pushing an object to move on a floor/table at each time step $t$ as
\begin{equation*}
\begin{split}
& E(t) = E_{trans} + E_{rot} = \mu m g  \Delta x + \mu \frac{m g}{X Y} \theta \Bigg[ \frac{X^3}{12} \\ & \left(\frac{\sin \phi}{\cos ^{2} \phi}  +\ln \frac{1+\sin (\phi)}{\cos (\phi)}\right) 
 - \frac{Y^3}{12} \left(\frac{-\cos \phi}{\sin ^{2} \phi}+\ln \left|\tan \frac{\phi}{2}\right|\right)\Bigg]
\end{split} 
\end{equation*}
where $m$ and $\mu$ are the object mass and the friction coefficient between the box and the floor/table. The contact surface has width $X$, length $Y$, with $\phi = \arctan \frac{Y}{X}$. The object displacement of time step $t$ is $\Delta x$ and the object rotates around its z-axis by an angle of $\theta$. All physical quantities involved here can be either obtained from the simulator or measured in the real-world irregardless of the robot platforms.

We then normalize $E(t)$ by its running bounds to get {\bf step physical cost} relative to all history steps the agent has taken so far, i.e. $c(t) = (E(t) - E_{min}(t)) / (E_{max}(t) - E_{min}(t))$. Similarly, for any successful trajectory $\tau$, we define its {\bf episode physical cost} as $c(\tau) = \frac{E(\tau) - E_{min}(\tau)}{E_{max}(\tau) - E_{min}(\tau)}$, which is normalized over the episode physical cost of all successful episodes so far. $c(t)$ and $c(\tau)$ will be used in our reward function to measure the physical expense of an action or action sequence.

\section{Physics-Aware Indoor Rearrangement Tasks} \label{sec:task}
We experiment in the 3D simulated environments built upon classroom scenes from OpenRooms dataset \cite{openrooms} by integrating Bullet \cite{bullet} physics engine and iGibson's renderer \cite{igibson}. Our tasks are performed by a simulated two-wheeled Fetch robot \cite{fetch} under joint space control with abilities other than locomotion disabled. The robot has a state space of 3D positions and orientations of itself and the interactive objects. It has four discrete actions: $[v,v]$ (move forward), $[0.5v, -0.5v]$ (turn left), $[-0.5v, 0.5v]$ (turn right) and $[0,0]$ (stop), where parameter $v$ is the maximum value of the robot's wheel angular velocity.
\vspace{-0.1cm}  
\subsection{Variable Friction Pushing Task} \label{two-band-task}
In this task, the agent and a box (10 kg) are initialized on a two-band floor. The red and blue bands have friction coefficients 0.2 and 0.8 respectively (Fig. \ref{fig:teaser}). The agent's objective is to push the box to the target rectangular region located on the other band along the most efficient trajectory without collision with any obstacles. This is taught by a physics-inspired reward function:
\begin{equation*} \label{eq:two_band_reward_with_energy}
\vspace{-0.1cm} 
r(\mathbf{s}_{t}, \mathbf{a}_{t}) = 
\begin{cases}
R & \text{succeed} \\
r_n & \text{agent collides with obstacles} \\ 
r_p \cdot c(t) & \text{agent pushes box} \\
r_e & \text{otherwise}
\end{cases}
\vspace{-0.1cm} 
\end{equation*}
where $R = 10 $ and $r_n = -10, r_e = -1, r_p = -0.5 \cdot c(t)$ in practice.
By weighting the constant pushing reward $r_p$ with step physical cost $c(t) \in [0,1]$, the agent tends to find the most physically efficient way rather than the fastest way.
\vspace{-0.1cm}  
\subsection{Variable Mass Pushing Task} \label{two-box-task}
In this task, the agent and two boxes are initialized in a circular region. The two boxes have the same shape, size and distance to the agent, but with different materials and weights (10 kg and 50 kg respectively) (Fig. \ref{fig:teaser}). The objective of the agent is to push at least one box outside the circle. With the physics-inspired reward defined below, the agent is learning to choose a lighter box to push: 
\begin{equation*} \label{eq:two_box_reward_with_energy}
r(\mathbf{s}_{t}, \mathbf{a}_{t}) = 
\begin{cases}
R \cdot (1-c(\tau)) & \text{succeed} \\
r_n & \text{agent collides with obstacles} \\ 
0 & \text{otherwise (time elapse)}
\end{cases}
\end{equation*}
where $R = 100$, $r_n=-10$ in practice. Weighting the succeed reward with the episode physical cost $c(\tau) \in [0,1]$ informs the agent to choose a policy succeeding in task at the minimum physical cost. 
\vspace{-0.1cm}  

\section{Experiments and Results} \label{sec:result}
\begin{figure}[!!t]
    \begin{minipage}[c]{0.49\columnwidth}
        \centering
        \includegraphics[width=\columnwidth]{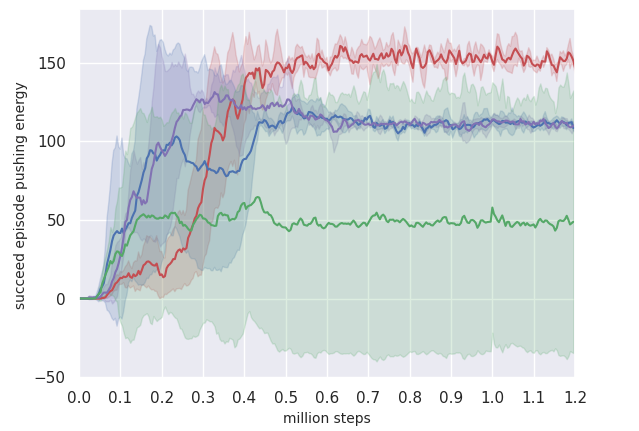}
    \end{minipage}
    \begin{minipage}[c]{0.49\columnwidth}
        \centering
        \includegraphics[width=\columnwidth]{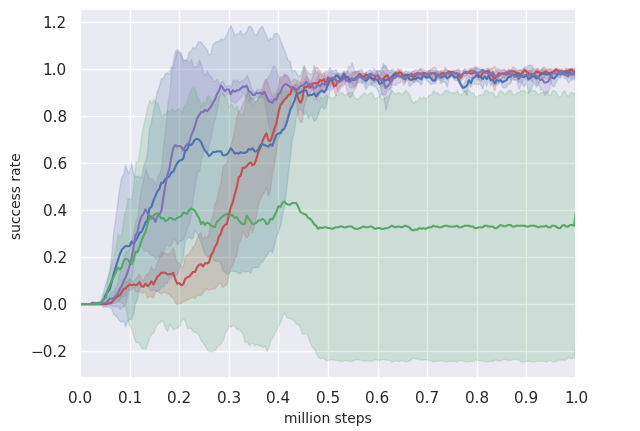}
    \end{minipage}
    \caption{Training curves of PPO on variable friction pushing tasks across 3 random seeds. Left: energy cost of successful episodes. Right: success rate. Red (No-physical-cost): policy trained with $c(t)=1$. Blue (Ours): policy trained with pushing energy normalized by running bounds. Magenta: policy trained with pushing energy normalized by fixed bounds. Green: policy trained with robot energy normalized by running bounds.} \label{fig:two_band_learning_curve}
\end{figure}

\begin{figure}
    \begin{minipage}[c]{0.49\columnwidth}
        \centering
        \includegraphics[width=\columnwidth]{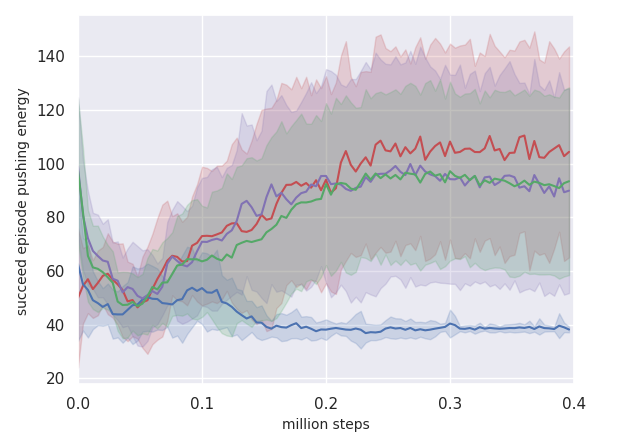}
    \end{minipage}
    \begin{minipage}[c]{0.49\columnwidth}
        \centering
        \includegraphics[width=\columnwidth]{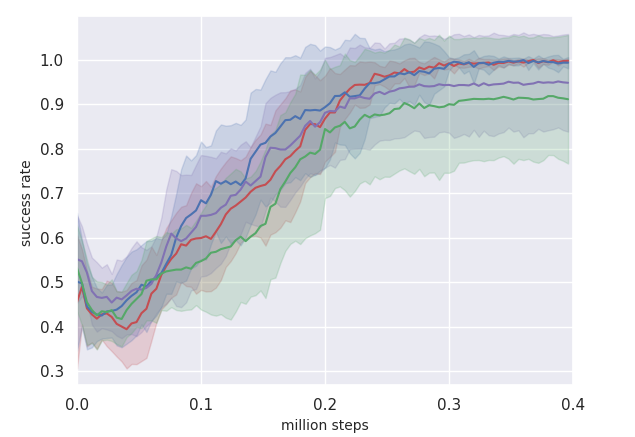}
    \end{minipage}
    \caption{Training curves of PPO on variable mass pushing tasks across 10 random configurations. Left: energy cost of successful episodes. Right: success rate. Red (No-physical-cost): policy trained with $c(\tau)=0$. Blue (Ours): policy trained with pushing energy normalized by running bounds. Magenta: policy trained with pushing energy normalized by fixed bounds. Green: policy trained with robot energy normalized by running bounds.} \label{fig:two_box_learning_curve}
\vspace{-0.5cm}  
\end{figure}

We trained the agents using proximal policy optimization (PPO) \cite{ppo} algorithms and evaluated the learned policies on the proposed tasks to answer the following questions:

{\bf Does our physics-inspired reward function correctly guide the agent to learn a physically efficient policy?} We compare the policies learned with (Blue) and without (Red) the physical cost in Fig.~\ref{fig:two_band_learning_curve} and \ref{fig:two_box_learning_curve}. The policies learned under the physics-inspired reward function has a clear drop on the mean energy cost of the successful episodes while converges to same near optimal success rate.  

\begin{figure}
    \begin{minipage}[l]{0.47\columnwidth} 
        \centering 
        \includegraphics[width=\columnwidth]{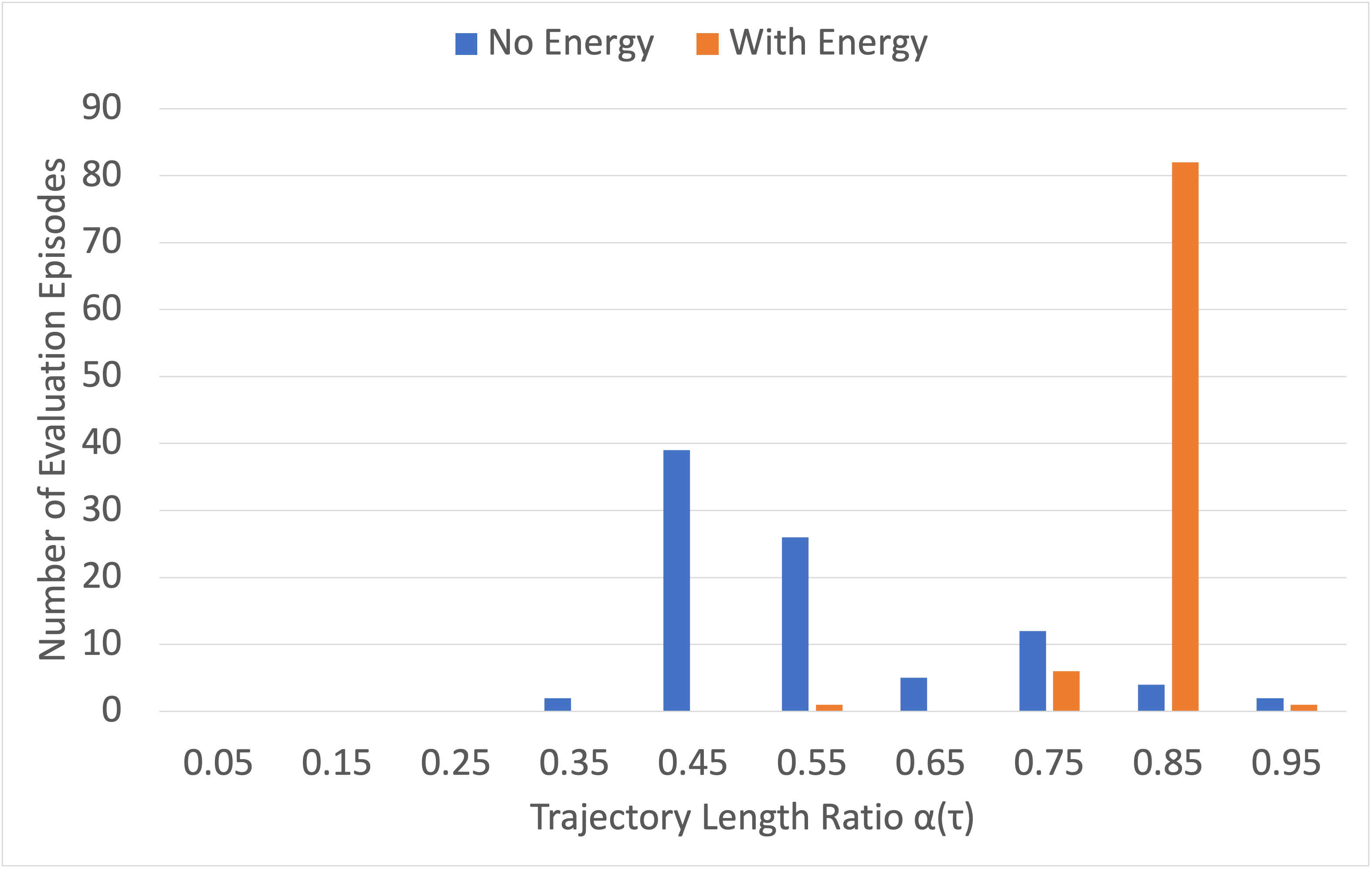}
        \caption{Frequency distributions of trajectory length ratio $\alpha(\tau) = low-friction-length / total-length$ of 30 evaluation trajectories. Blue: policy trained without physical cost. Orange: policy trained with physical cost.}
        \label{fig:two_band_dist}
    \end{minipage}
    \hfill
    \begin{minipage}[r]{0.47\columnwidth} 
        \centering 
        \includegraphics[width=\columnwidth]{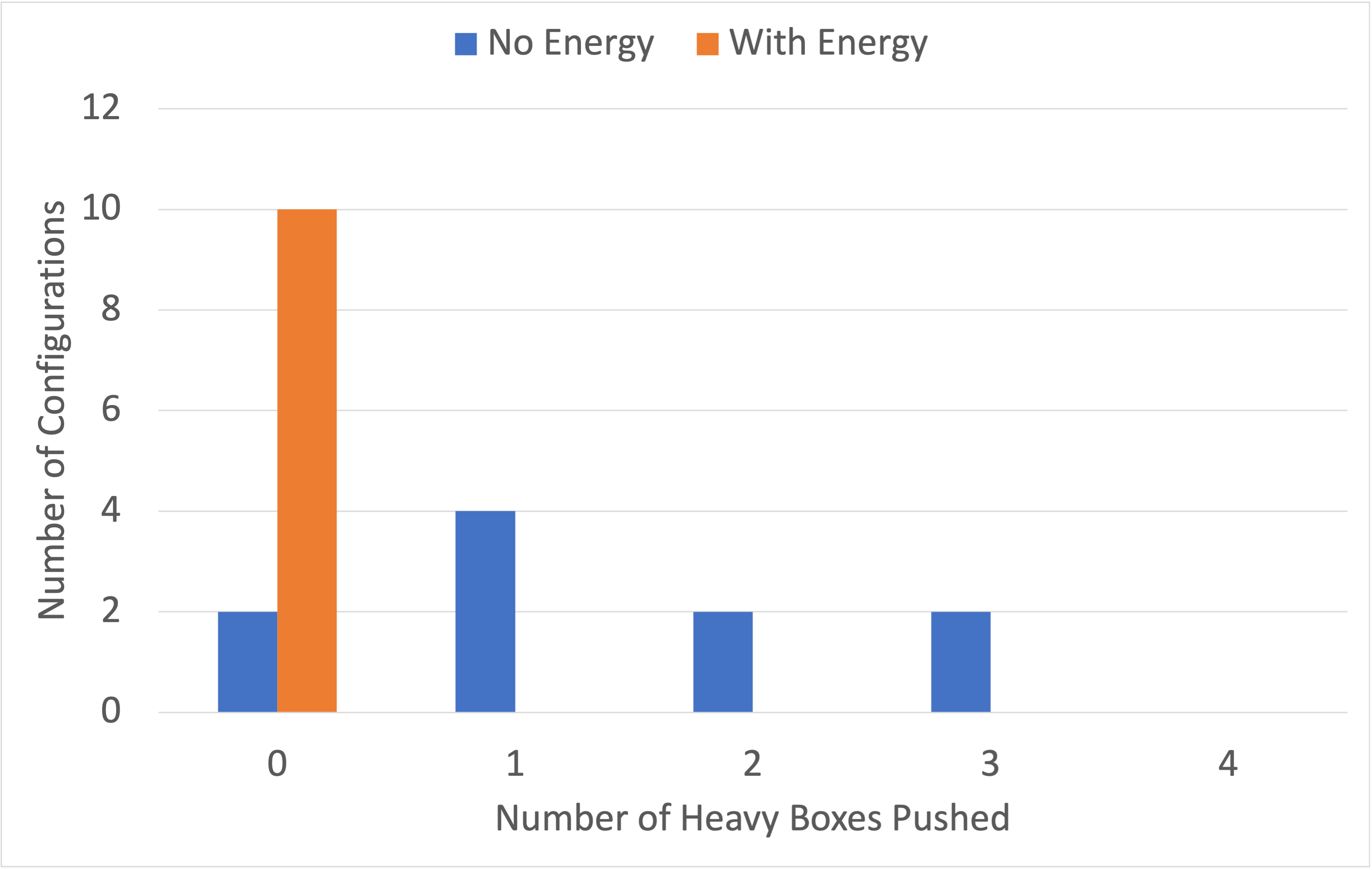}
        \caption{Frequency distributions of the number of times the heavy boxes are chosen to push over 10 random configurations (from 0 to 4). Blue: policy trained without physical cost. Orange: policy trained with physical cost.}
        \label{fig:two_box_dist}
    \end{minipage}
\vspace{-0.5cm}    
\end{figure}

{\bf Does the policy learned under physics-inspired reward function reflect the agent's awareness of mass and friction?} We examine the distributions of evaluation trajectories on the preferences over low vs. high friction in Fig.~\ref{fig:two_band_dist} and light vs. heavy box in Fig.~\ref{fig:two_box_dist} with or without physical cost. The policy learned with physical cost has shown a clear preference while the counterpart without physical cost made the decision with nearly equal probabilities.

{\bf How does our definition of physical cost compare to other design strategies on learning successful and efficient policies?} We ablate our physical cost to understand the role of the following factors: (1) Using pushing energy instead of robot output energy. (2) Normalizing the cost by running bounds instead of fixed bounds, which are estimated according to extra prior knowledge about the environment and the task. The comparisons of the training curves over the combinations of these factors are shown in Fig.~\ref{fig:two_band_learning_curve} and \ref{fig:two_box_learning_curve}. We observe that in most cases, removing either of these factors will lead to a drop in success rate or physical efficiency.

\section{Future work} \label{sec:conclusion}
Our work made a step towards learning physics-aware policies which are responsive to different masses and frictions. Potential future works include: (1) Extending the method to vision-based RL for better generalization of the physical risk. (2) Transferring the physics-inspired policies, value functions, experiences to downstream safe-critic tasks to avoid risky behaviors such as falling down on slippery floor, pushing unmovable objects, etc.

\section*{Acknowledgments}
We thank NSF CAREER 1751365, CHASE-CI and generous gifts from Adobe, Google and Qualcomm.

\bibliographystyle{IEEEtran.bst}
\bibliography{egbib}

\end{document}